\documentclass[aip,reprint]{revtex4-1}

\usepackage{amsmath}
\usepackage{lineno}
\usepackage{graphicx}
\usepackage{subfigure}
\usepackage{bm}
\usepackage{natbib}
\usepackage{float}
\usepackage[ruled]{algorithm2e}
\usepackage{color}

\definecolor{Red}{rgb}{1,0,0}  
\usepackage{afterpage}

\draft % marks overfull lines with a black rule on the right

\begin{document}

%\preprint{}
%\pagewiselinenumbers
%\switchlinenumbers
%\linenumbers 
\title{Short-term Load Forecasting Based on Hybrid Strategy Using Warm-start Gradient Tree Boosting} %Title of paper



\author{Yuexin Zhang}
\email{230169211@seu.edu.cn}
\altaffiliation{}
\affiliation{School of Instrument Science and Engineering, Southeast University, Nanjing 210096, China.}
\affiliation{Department of Electrical and Computer Engineering, National University of Singapore, 117576, Singapore.}

\author{Jiahong Wang}
\email{jiahong4@illinois.edu}
\altaffiliation{}
\affiliation{Department of Electrical and Computer Engineering, University of Illinois at Urbana-Champaign, IL 61801, United Sates.}


\date{\today}

\begin{abstract}
A deep-learning-based hybrid strategy for short-term load forecasting is presented. The strategy proposes a novel tree-based ensemble method Warm-start Gradient Tree Boosting (WGTB). Current strategies either ensemble submodels of a single type, which fail to take advantage of the statistical strengths of different inference models. Or they simply sum the outputs from completely different inference models, which doesn't maximize the potential of ensemble. Inspired by the bias-variance trade-off, WGTB is proposed and tailored to the great disparity among different inference models on accuracy, volatility and linearity. The complete strategy integrates four different inference models of different capacities. WGTB then ensembles their outputs by a warm-start and a hybrid of bagging and boosting, which lowers bias and variance concurrently. It is validated on two real datasets from State Grid Corporation of China of hourly resolution. The result demonstrates the effectiveness of the proposed strategy that hybridizes the statistical strengths of both low-bias and low-variance inference models.
\end{abstract}


\maketitle %\maketitle must follow title, authors, abstract and \pacs

%\section{}
%\label{}
%\subsection{}
%\subsubsection{}

% If in two-column mode, this environment will change to single-column format so that long equations can be displayed. 
%\begin{widetext}
%$$\mbox{put long equation here}$$
%\end{widetext}

% Figures should be put into the text as floats. 
% Use the graphics or graphicx packages (distributed with LaTeX2e).
% Put the label that you will use with \ref{} command in the braces of the \label{} command.
%
% \begin{figure}
% \end{figure}

% \begin{table}
% \end{tabular}
% \end{table}

%\begin{acknowledgments}
% Put your acknowledgments here.
%\end{acknowledgments}

%\bibliography{your-bib-file}

\section{Introduction}
\label{S:1}
Electric power is a non-storable product, electric power utilities need ensure a precise balance between the electricity production and consumption at any moment. Therefore, load forecasting plays a vital role in the daily operational management of power utility, such as energy transfer scheduling, unit commitment, load dispatch, and so on \citep{intro_1_1,intro_1_3,intro_1_2}. With the emergence of load management strategies, it is highly desirable to develop accurate load forecasting models for these electric utilities to achieve the purposes of higher reliability and management efficiency \citep{intro_1_4,intro_1,intro_2}.

The inference models for Short-term Load Forecasting (STLF) can be classified into two categories\citep{intro_3}. The first type is autoregressive models (or linear models). Due to their small capacity, they often have small variances but high biases. One of the most common forecasting techniques amongst the autoregressive models is the Autoregressive Integrated Moving Average (ARIMA) model \citep{intro_4_0, intro_4}. The other type is machine learning models (or nonlinear models). Because of their greater capacity, they have relatively smaller bias but higher variance. The Support Vector Regression (SVR) is a highly effective model in machine learning and has the capability of solving nonlinear problems, even with small quantities of training data \citep{intro_5,intro_7, intro_9}. As a fresh field of machine learning, Artificial Neural Network (ANN) has attracted much more attention for STLF \citep{intro_12, intro_13, intro_17_2}, especially the Long Short-term Memory (LSTM) neural network \citep{intro_16,intro_17}. The SVR and ANN models usually require a substantial amount of time to train the forecasting model. The Extreme Learning Machine (ELM) can reduce the training time of neural networks and achieve a good accuracy of forecasting. Thus ELM has become a popular forecasting technique for STLF due to its faster performance \citep{intro_14_2, intro_15, intro_15_2}.
    
Nonetheless, all inference methods have their flaws and weaknesses, which, if they are used singly, will impinge on the forecasting precision. Furthermore, most of the forecasting methods rely heavily on the presumed data patterns so that no single model is suitable for all. Hybrid models are thus created to aggregate the advantages of individual models \citep{intro_17_3, intro_17_4}. Hybrid models can be classified into two categories based on the number of types of inference submodels. Models in the first category contain a single inference submodel, usually ANN. They sum the outputs from multiple identical ANNs using bagging \citep{intro_21_4}, boosting \citep{intro_21_5} or a combination of both \citep{intro_21_6}. And a fuzzy logic based approach was recently proposed by Sideratos that ensembles multiple radial basis function-convolutional neural networks \citep{intro_21_2}. Models in the other category consist of multiple types of inference submodels, such as ARIMA and SVRs \citep{intro_18,intro_19}, ARIMA and ANN \citep{intro_22}, ANN, SVR and Gaussian process regression \citep{intro_20}, ELM, phase space reconstruction and a least squares support vector machine \citep{intro_21}.

Whereas the aforementioned hybrid strategies could generate decently accurate and reliable load prediction, they are still defective in some ways. The first type of hybrid model contains only a single inference model and fails to take advantage of the statistical strengths of different inference models. While the second type of hybrid model does make the use of both autoregressive models and machine learning models, they do not fully consider the bias-variance trade-off. The methods they use are essentially a simple summation either a weighted sum or iterative sum. There is much room for improvement in balancing the great bias/variance disparity among different inference models, which maximizes the potential of the ensemble.

To overcome the shortcomings of existing approaches, we propose a novel hybrid strategy inspired by the bias-variance trade-off. The strategy incorporates the following novelty points:
\begin{enumerate}
    \item First, the inference submodels are selected based on bias and variance. While low-bias models (e.g. LSTM) give accurate predictions, they suffer from great variance. Low-variance models (e.g. ARIMA) are hybridized to balance the former. The hybrid of both low-bias and low-variance models improves the accuracy of the predictions.
    \item Second, a novel ensemble model Warm-start Gradient Tree Boosting (WGTB) is proposed to lower the variance of low-bias models by ensembling them with low-variance but high-bias models, such that bias and variance are reduced concurrently. The WGTB outperforms simple summations and existing ensemble models.
    \item The novelties of WGTB are reflected in its initialization and convergence that lower bias and variance respectively. The low bias is achieved by a warm-start initialization by ElasticNet. ElasticNet allows WGTB to lower the bias by picking its initialization value sparsely from all inference submodels and avoid high-bias models automatically. The low variance is achieved by a boosting-bagging algorithm that fits a bag of regression trees iteratively. Regression trees utilize the outputs of low-variance models to lower the variance of high-variance models. And the integration of bagging in boosting further allows the variance to be reduced to the full extent. 
    \item WGTB is a tree-based ensemble model for STLF. To the best of authors’ knowledge, tree-based ensemble models have never been used in deep load forecasting models. The insusceptibility of regression trees to the bias of submodels allows them to lower the variance without being affected by the bias. This feature is unachievable without the tree-based ensemble models.
    \item Rigorous comparative experiments are carried out, and the results demonstrate that the hybrid model has a better forecasting accuracy than individual submodels, and the ensemble model WGTB performs better than existing ensemble models.
\end{enumerate}

The rest of the paper is organized as follows. Section \ref{sec:overview} presents the framework of hybrid strategy and the novel ensemble algorithm WGTB. Section \ref{sec:results} analyzes the prediction results of the proposed model. Finally, conclusions are given in Section \ref{sec:conclusion}.

\section{The Proposed Hybrid Strategy for STLF}\label{sec:overview}
The hybrid strategy is inspired by bias-variance trade-off in the learning theory. For squared error, the bias-variance decomposition is formulated as the following way. Assume we use random sampling scheme to sample a fixed-size training set $D=\{(x_1,y_1),...,(x_N,y_N)\}$. And we denote the prediction of a deterministic model trained by training set $D$ as $f(x;D)$. For any trained model, we can get the same generalization error decomposition as Equ. \ref{eqn:err_decompos} \cite{extratree},
\begin{linenomath*}\begin{align}
    E_D[E_{Y}[(Y,f(x;D))^2]]&=\sigma^2(x)+{Bias}_D(x)^2+{Var}_D(x)\label{eqn:err_decompos}
\end{align}\end{linenomath*}
where
\begin{linenomath*}\begin{align}
    \sigma^2(x)&=E_{Y|x}[(Y-f_B(x))^2]\\
    Bias_{D}(x)^2&=(f_B(x)-\overline{f}(x))^2\\
    {Var}_D(x)&=E_{D}[(f(x;D)-\overline{f}(x))^2]
\end{align}\end{linenomath*}
with
\begin{linenomath*}\begin{align}
    f_{B}(x)&=E_{Y|x}[Y]\\
    \overline{f}(x)&=E_{D}[f(x;D)]
\end{align}\end{linenomath*}
The first term of the decomposition $\sigma^2$ is the irreducible error, which is the theoretical lower bound. The second term ${Bias}_D(x)^2$ is the squared bias of the model, measuring the discrepancy between Bayes optimal model $f_B(x)$ and the average model $\overline{f}(x)$. And the third term is the variance which indicates how far the model deviates from its mean due to the sampling of training set.

Complex nonlinear models such as LSTM have sophisticated assumptions. Therefore, they are flexible to adapt to a variety of underlying probability distributions and have small theoretical biases on average. But in practice, they often suffer from great variance, and we cannot obtain a model close to the average. On the other hand, low-variance models such as the linear ARIMA model are limited by their simplifying assumptions which lead to great biases. So these models are often discarded and replaced by the former low-bias models.

\begin{figure}[!hbt] 
%\vspace*{-2.5cm}
\centering
\includegraphics[scale=0.5]{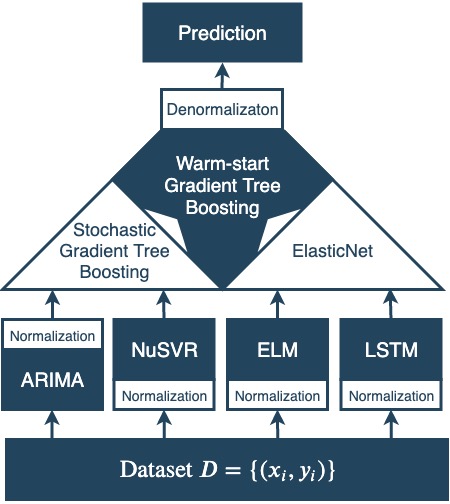}
\caption{Hybrid strategy for STLF}
\label{fig:hybridmodel}
\end{figure}

A hybrid strategy is thus proposed to lower the variance of low-bias models using low-variance but high-bias models. The hybrid strategy integrates models of different capacity using a novel ensemble model WGTB. The framework of the hybrid strategy is presented in Fig. \ref{fig:hybridmodel}. The load dataset is first pre-processed. Then four submodels (i.e.,ARIMA, NuSVR, ELM and LSTM) with various bias and variance are used to prediction. Thirdly, the proposed ensemble model WGTB forecasts the load based on the juxtaposition of outputs from four submodels. Finally, accurate load forecasts are obtained after data post-processing.

An additional benefit from the hybrid strategy is that the input dimension to the ensemble model WGTB is independent of the high dimensionality and the high volume of input data. Submodels (e.g., LSTM) are very capable of dimensional reduction, so the input dimension to WGTB is only directly proportional to the number of submodels.

The rest of the section first revisits stochastic gradient tree boosting (SGTB) and then describes how warm-start gradient tree boosting builds upon SGTB to achieve both low bias and low variance. Finally, the section gives a brief talk on the four inference models, in which the architecture of our LSTM neural network model is described.

\subsection{stochastic gradient tree boosting}
SGTB was first introduced by Friedman \citep{gbrt_random} who added random sampling technique to original gradient tree boosting \citep{gbrt_invent}. The model is a generalized greedy function approximator using boosting method. Each weak regressor in the model is a decision tree, namely Classification And Regression Trees (CART). The mathematical formulation is in additive fashion,
\begin{linenomath*}\begin{equation}
    F({x})=\gamma_0+\sum_{m=1}^{M}\gamma_m h_m({x})\label{eqn:gtb_add}
\end{equation}\end{linenomath*}
where 
${x}$ is input vector, $M$ is the number of decision trees,
$h_m$ is the $m^{th}$ decision tree,
and $\gamma_m$ is weight coefficient for decision tree $h_m$.

The loss objective of the model is,
\begin{linenomath*}\begin{equation}
    \text {min} \ E_{y,{x}}[L(y,{ x})]:=\sum_{i=0}^{N-1} L(y_i, F({ x}_i))\label{eqn:gtb_obj}
\end{equation}\end{linenomath*}
where $L$ is mean square error loss function, $({x}_i,y_i)\in D$ is a sample in the dataset $D$, $x$, $y$
denote arbitrary input and groundtruth output respectively, and $N$ is the number of samples in dataset.

Gradient tree boosting builds its model in iterative method as shown below,
\begin{linenomath*}\begin{align}
    F_0(x)&=\gamma_0\label{eqn:gtb_zero}\\
    F_m(x)&=F_{m-1}({x})+\gamma_m h_m({ x})\quad\forall m>0\label{eqn:gtb_iter}
\end{align}\end{linenomath*}

The objective for $m^{th}$ iteration then becomes
\begin{linenomath*}\begin{equation}
   \text {min}\ \sum_{i=0}^{N-1} L(y_i, F_m({ x}_i)):=\sum_{i=0}^{N-1} L(y_i, h_m({ x}_i)+F_{m-1}({ x}_i))\label{eqn:gtb_iterobj}
\end{equation}\end{linenomath*}

To optimize the objective in Equ. \ref{eqn:gtb_iterobj}, the algorithm first optimizes $h_m$ by setting $\gamma_m$ to 1. Then it optimizes $\gamma_m$ by fixing $h_m$. To find $h_m$, gradient tree boosting uses steepest descent algorithm,
\begin{linenomath*}\begin{equation}
    F_m({x})=F_{m-1}({ x})-\gamma_m\nabla_F\sum_{i=0}^{N-1}L(y_i,F_{m-1}({ x}_i))\label{eqn:gtb_grad}
\end{equation}\end{linenomath*}

So in each iteration, a new decision tree is fitted to predict the additive inverse of gradient,
\begin{linenomath*}\begin{equation}
    h_m(x)\gets-\nabla_F\sum_{i=0}^{N-1}L(y_i,F_{m-1}({ x}_i))\label{eqn:gtb_hm}
\end{equation}\end{linenomath*}

After $h_m$ is determined, the optimization of $\gamma$ is shown as follows,
\begin{linenomath*}\begin{align}
    \gamma_0&=\mathop{\rm {argmin}}_{\gamma}\sum_{i=0}^{N-1}L(y_i,\gamma)\label{eqn:gtb_gamma0}\\
    \gamma_m&=\mathop{\rm {argmin}}_{\gamma}\sum_{i=0}^{N-1}L(y_i,\gamma h_m({ x}_i)+F_{m-1}({ x}_i))\quad\forall m>0\label{eqn:gtb_gamma}
\end{align}\end{linenomath*}

Two techniques are also developed together with gradient tree boosting, first of which is shrinkage, an effective technique in controlling learning rate. The regularization through the shrinkage parameter $\nu$ and the number of components $M$ can prevent overfitting, and the former provides superior results. Smaller values of $\nu$ can give rise to larger optimal $M$-values, thus providing higher accuracy, but there is a diminishing return for the smallest values. And increasing the size of $M$ produces a proportionate increase in computation. The best value for $\nu$ depends on $M$. The latter should be made as large as is computationally convenient or feasible. The value of $\nu$ should then be adjusted so that the lack of fit achieves its minimum close to the value chosen for $M$. The shrinkage parameter $\nu\leq0.1$ can lead to better generalization error. Through our experiments, the parameters $\nu$ and $M$ are set to 0.05 and 70, respectively. Then the Equ. \ref{eqn:gtb_iter} becomes,
\begin{linenomath*}\begin{equation}
    F_m(x)=F_{m-1}({x})+\nu\gamma_m h_m({x})\quad\forall m>0\label{eqn:gtb_iter_shrink}
\end{equation}\end{linenomath*}

The other technique is random sampling. During each iteration, instead of training the base decision tree with the entire dataset, a subsample of training data is drawn at random (without replacement) and it is used to fit the base decision tree. Using the subsample causes the variance of the individual base learner estimates at each iteration to increase. However, there is less correlation between these estimates at different iterations. This tends to reduce the variance of the combined model (7), which, in effect, averages the base learner estimates. The latter averaging effect dominates the former one even for surprisingly small subsamples. Thus the variance is reduced, and approximation accuracy and execution speed of gradient boosting is substantially improved.

The stochastic gradient tree boosting algorithm can be summarized as,\\
\begin{algorithm}[H]
\SetAlgoLined
\KwData{Dataset $D=\{({x}_i,y_i)\}^N$, shrinkage parameter $\nu$, number of decision trees $M$, subsample ratio $\eta$}
\KwResult{Weight coefficient $\gamma_m$ and decision tree $h_m$}
 $N^{'}\gets Floor(\eta N)$\;
 $\gamma_0\gets\mathop{argmin}_{\gamma}\sum_{i=0}^{N-1}L(y_i,\gamma)$\;
 \For{$m \gets 1$ {to} $M$}
 {
 $\{({ x}^{'}_j,y^{'}_j)\}^{N^{'}}\gets Random\_Subsample(\{({ x}_i,y_i)\}^N)$\;
 $r_{jm}\gets-\nabla_F L(y^{'}_j,F_{m-1}({ x}^{'}_j))\quad\forall j$\;
 Fit decision tree $h_m({ x}^{'})$ with $\{({ x}^{'}_j,r_{jm})\}^{N^{'}}\quad\forall j$\;
 $\gamma_m\gets\mathop{argmin}_{\gamma}\sum_{j=0}^{N^{'}-1}L(y^{'}_j,\gamma h_m({ x}^{'}_j)+F_{m-1}({ x}^{'}_j))$\;
 $F_m({ x})\gets F_{m-1}({ x})+\nu\gamma_m h_m({ x})$;
 }
 \caption{Stochastic Gradient Tree Boosting\label{alg:sgtb}}
\end{algorithm}

\subsection{Warm-start gradient tree boosting}
WGTB is designed to lower both (squared) bias term $Bias_D(x)^2$ and variance term $Var_D(x)$ in Equ. \ref{eqn:err_decompos}. WGTB achieves the two objectives separately by improving the initialization and convergence of SGTB. While WGTB is named to be an improvement on SGTB, they are completely different in the underlying heuristics. For one thing, while SGTB is purely boosting algorithm, WGTB hybridizes bagging and boosting. For two thing, the objective of boosting in SGTB is to lower the bias while the objective of boosting in WGTB is to lower the variance. The initialized value in SGTB is an average, and SGTB lowers its bias iteratively by fitting decision trees. On the other hand, the initialized value in WGTB is a low-bias prediction already. WGTB uses boosting to lower its variance by utilizing predictions from low-variance inference models. 

\subsubsection{Initialization -- low bias}
The initialized value for the iterative algorithm is a crucial factor in lowering the bias. The proposed WGTB differs from SGTB in a linear warm start. More specifically, SGTB uses a constant value, denoted as $F_0({ x})=E_Y[Y]$, as its initialized value. The value is the average (expectation) of $Y$ in the training set. The choice of this value assumes that SGTB has no knowledge of $Y$, and assumes $Y$ is drawn from a uniform distribution. So SGTB doesn't utilize the outputs of individual inference models to lower the bias the initialized value. By contrast, WGTB introduces a likelihood function $P(Y|X)$ that is conditioned on $X$ and corresponds to a normal distribution, in which $X$ are the outputs of individual inference models. The likelihood function implies that $F_0({ x})$ is modeled as linear regression. To prevent the high-bias model to impedes on the performance, both L1 and L2 regularization terms are added. This choice coincides with ElasticNet, which is one of the most generalized versions of linear regressions that include both L1 and L2 regularization terms of the coefficients, which allows it to learn a sparse model while maintaining regularization properties. Equ. \ref{eqn:gltb_elasticnet} is the training of ElasticNet, and $F_0({ x})$ is its predicted value. Mathematically, the equations become
\begin{linenomath*}\begin{equation}
    F_0({ x})=\gamma_0={x}^T{ w}-y\label{eqn:gltb_F0}\\
   \end{equation}\end{linenomath*}
\begin{linenomath*}\begin{equation}
    { w}=\mathop{\rm {argmin}}_{ w}\ E_{{ x},y}{ [}\ \frac{1}{2 N}\Vert { x}^T{ w}-y\Vert^2_2\ {]} + \alpha\rho\Vert { w}\Vert_1 + \alpha(1-\rho)\Vert {w}\Vert_2^2\label{eqn:gltb_elasticnet}
\end{equation}\end{linenomath*}
where $\alpha$ stands for regularization coefficient of ElasticNet,
and $\rho$ stands for L1 ratio of ElasticNet.

The ElasticNet model doesn't seem to add much from the average (expectation) value used by SGTB. But in fact, its output is generated by ElasticNet and individual inference models (e.g. LSTM and ELM) as a whole. The capacity of the entire architecture is much larger than ElasticNet and any individual inference model, which accordingly reduces the bias.

In practice, the choice of hyperparameters of ElasticNet has a great impact on the number of total iterations. Exhaustive search is a popular method in tuning hyperparameters. It is incorporated into the model to search for the best ElasticNet in the warm start. Since there are only two hyperparameters in ElasticNet ($\alpha$ and $\rho$), the exhaustive search is quick enough. Empirically, a guess of 5$\sim$6 $\alpha_s$ and 4$\sim$5 $\rho_s$ usually results in a decent outcome.

\subsubsection{Convergence -- low variance}
Unlike SGTB, the initialized value $F_0({ x})$ of WGTB is already a low-bias prediction. Intuitively, the initialized value $F_0({ x})$ is a weighted average of multiple low-bias inference models. If one view ElasticNet and these low-bias inference models as a whole, the capacity of the architecture is greater than any single inference model. Therefore, the initialized value comes with a large variance. To lower its variance, WGTB corrects the error by fitting base tree estimators iteratively. Equ. \ref{eqn:gltb_additive_F} shows its mathematical formulation. 
\begin{linenomath*}\begin{equation}
    F({ x})=f({ x})+\sum_{m=1}^{M}e_m({ x})\label{eqn:gltb_additive_F}
\end{equation}\end{linenomath*}
where 
$f({ x})$ is ElasticNet model, and
$e_m({ x})\ \forall m$ is the base tree estimator. The base tree estimator $e_m({ x})$ is selected as ExtraTree (Extremely Randomized Tree), and the detailed reason is given in the last paragraph in this Section.

The greatest difference between SGTB and WGTB is in the use of bagging. SGTB is solely a boosting algorithm. Its base tree estimator is CART, since its aim is to reduce bias. On the other hand, WGTB fits a bag of regression trees. Bagging is known for reducing the variance. To reduce the variance as far as possible, random sampling is introduced. The rationale for random sampling and bagging is formulated mathematically in the following way. Assume we still use random sampling scheme to sample a fixed-size training set $D=\{(x_1,y_1),...,(x_N,y_N)\}$. But we add an random variable $\epsilon$ to the model such that individual randomized regression tree is denoted as $f_r(x;D,\epsilon)$. The generalization error decomposition is then changed from Equ. \ref{eqn:err_decompos} to Equ. \ref{eqn:err_decompose_espilon}\cite{extratree}.
\begin{linenomath*}\begin{equation}
\begin{aligned}
    E_D[E_{Y}[(Y,f_r(x;D,\epsilon))^2]]&=\sigma^2(x)+(f_B(x)-\overline{f_r}(x))^2\\
    &+{Var}_{D,\epsilon}(f_r(x;D,\epsilon))
\end{aligned}
\label{eqn:err_decompose_espilon}
\end{equation}\end{linenomath*}
where
\begin{linenomath*}\begin{align}
    \overline{f_r}(x)&=E_{D,\epsilon}[f_r(x;D,\epsilon)]\\
    \sigma^2(x)&=E_{Y|x}[(Y-f_B(x))^2]\\
    {Var}_{D,\epsilon}(f_r(x;D,\epsilon))&=E_{D,\epsilon}[(f_r(x;D,\epsilon)-\overline{f_r}(x))^2]\label{eqn:var_epsilon}
\end{align}\end{linenomath*}
The Equ. \ref{eqn:var_epsilon} is further decomposed to two positive terms according to the law of total variance:
\begin{linenomath*}\begin{equation}
\begin{aligned}
{Var}_{D,\epsilon}(f_r(x;D,\epsilon))&={Var}_{D}(E_{\epsilon|D}[f_r(x;D,\epsilon)])\\
&+E_{D}[{Var}_{\epsilon|D}(f_r(x;D,\epsilon))]
\end{aligned}
\end{equation}\end{linenomath*}
Each regression tree differs from others by having different $\epsilon$ value. Assume there are $M$ trees and ${\epsilon}_i$ are drawn independently from the same distribution $P(\epsilon|D)$. When bagging averages the outputs of regression trees, the variance is lowered according to the size of $M$ as it is shown in Equ. \ref{eqn:avg_var}, and bias remains the same value under the same assumption as it is shown in Equ. \ref{eqn:avg_bias}.
\begin{linenomath*}\begin{equation}
\begin{aligned}
    {Var}_{D,\epsilon^M}(f_{a^M}(x;D,\epsilon^M))&={Var}_{D}(E_{\epsilon|D}[f_r(x;D,\epsilon)])\\
    &+E_{D}[{Var}_{\epsilon|D}(f_r(x;D,\epsilon))]/M
\end{aligned}
\label{eqn:avg_var}
\end{equation}\end{linenomath*}
\begin{linenomath*}\begin{equation}
\begin{aligned}
    \overline{f_{a^M}}(x)&=E_{D,\epsilon^M}[f_{a^M}(x;D,{\epsilon^M})]\\
    &=\frac{1}{M}\sum_{i=1}^{M}E_{D,\epsilon_i}[f_r(x;D,\epsilon_i)]\\
    &=\overline{f_r}(x)
\end{aligned}
\label{eqn:avg_bias}
\end{equation}\end{linenomath*}
To model the bag of randomized regression trees, there are plenty of existing algorithms. The bagging of WGTB is carefully chosen to be modeled by ExtraTree (Extremely Randomized Tree)\citep{extratree}. Compared to other existing randomized bagging algorithm such as Random Forest, ExtraTree introduces the most randomness to individual regression tree. This allows the base tree estimator of WGTB to have the lowest possible variance. The complete pseudocode for WGTB is finally presented as

\begin{algorithm}[H]
\SetAlgoLined
\KwData{Dataset $D=\{({x}_i,y_i)\}^N$, shrinkage parameter $\nu$, number of decision trees $M$, subsample ratio $\eta$, regularization coefficient of ElasticNet $\alpha=\{\alpha_a\}^A$, L1 ratio of ElasticNet $\rho=\{\rho_h\}^H$}
\KwResult{Weight coefficient $\gamma_m$ and decision tree $h_m$}
 $N^{'}\gets Floor(\eta N)$\;
 ${ w}\gets\mathop{argmin}_{ w}\ E_{{ x},y}{ [}\ \frac{1}{2 N}\Vert {x}^T{ w}-y\Vert^2_2\ { ]} + \alpha_a\rho_h\Vert { w}\Vert_1 + \alpha_a(1-\rho_h)\Vert { w}\Vert_2^2\quad\forall \alpha_a\in\alpha,\rho_h\in\rho$\;
 $\gamma_0\gets{ x}^T{ w}-y$\;
 \For{$m \gets 1$ {to} $M$}
 {
 $\{({x}^{'}_j,y^{'}_j)\}^{N^{'}}\gets Random\_Subsample(\{({ x}_i,y_i)\}^N)$\;
 $r_{jm}\gets-\nabla_F L(y^{'}_j,F_{m-1}({ x}^{'}_j))\quad\forall j$\;
 Fit ExtraTree $h_m({ x}^{'})$ with $\{({ x}^{'}_j,r_{jm})\}^{N^{'}}\quad\forall j$\;
 $\gamma_m\gets\mathop{argmin}_{\gamma}\sum_{j=0}^{N^{'}-1}L(y^{'}_j,\gamma h_m({ x}^{'}_j)+F_{m-1}({x}^{'}_j))$\;
 $F_m({ x})\gets F_{m-1}({ x})+\nu\gamma_m h_m({ x})$;
 }
 \caption{Warm-start Gradient Tree Boosting\label{alg:gltb_complete}}
\end{algorithm}

\subsection{Inference Submodels}
Four inference models of different variances and biases (i.e., auto-regressive integrated moving average, nu support vector regression, extreme learning machine and long short-term memory neural network) are selected for individual prediction before ensemble.
\subsubsection{Autoregressive integrated moving average}
ARIMA is a forecasting technique that projects the future values of a series based entirely on the inertia of the series itself  \citep{arima_apply1, arima_apply2}. The model is often written as ARIMA $(p,d,q)$. The parameters $p$, $d$, $q$ respectively denote the order of the autoregressive model, the degree of differencing and the order of the moving-average model \citep{arima_exp}. The equation of ARIMA is given as follows,
\begin{linenomath*}\begin{equation}
\label{eq:arima}
x_{t}=c+\sum_{i=1}^{p}\phi_{i}x_{t-i}+\epsilon_{t}+\sum_{i=0}^{q}\theta_{i}\epsilon_{t-i}
\end{equation}\end{linenomath*}
where $x_{t}$ is $d^{th}$ original time series, $c$ is constant term, $\phi_{i}$ is autocorrelation coefficient, $\theta_i$ is partial autocorrelation coefficient, and $\epsilon_t$ is corresponding error.

\subsubsection{Nu support vector regression}
To control the number of support vectors and training errors of traditional SVR, NuSVR adds a parameter $\nu$ to restrict the regularization coefficient, where $0 \leq \nu \leq 1$ \citep{nusvr}. The regression formula can be expressed as,
\begin{linenomath*}\begin{equation}
 f(x)=\sum_{i=1}^{N}w_i\phi_i(x)+b
\label{eqn:nusvr_eqn}
\end{equation}\end{linenomath*}
where $w_i$ is the coefficient, $\phi_i(x)$ is named feature, $N$ is the number of input data, and $b$ is the bias term.

The coefficients ${w_i}_{i=1}^N$ can be obtained by optimizing the following quadratic programming problem,
\begin{linenomath*}\begin{equation}
\mathop{\rm{min}}_w\ \Vert w\Vert^2_2 + \frac{1}{m \nu}L(x,f(x)) \label{eqn:nusvr_obj}
\end{equation}\end{linenomath*}
where $L$ is the loss function, and $m$ is corresponding regularization factor.
\subsubsection{Extreme learning machine}
ELM is known to be robust, highly accurate, and computationally efficient \citep{elm_apply1, elm_apply2}. Unlike traditional neural networks where weights of both layers are trainable, ELM is a special case of single-hidden-layer fully connected neural network. Only the weights of output layer are trainable and the weights of the hidden layer are randomly initialized and immutable during training. This benefits ELM to have a global optimum \citep{elm_huang}.

The output function $f_{L}({x}) $ of ELM with $L$ hidden neurons is
\begin{linenomath*}\begin{equation}
f_{L}({x}) = \sum_{i = 1}^{L}b_{i}(\sigma({x^Ta_i}) + e_i)
\label{eqn:elm_fl}
\end{equation}\end{linenomath*}
where ${x}=[x_1,...,x_d]^T$ is the $d$-dimensional input vector, ${ b}=[b_1,...,b_L]^T$ is the vector of output layer weights, ${a_i}=[a_{i1},...,a_{id}]^T$ is the weight vector of the $i^{th}$ neuron in hidden layer, $e_i$ is the bias of the $i^{th}$ neuron in hidden layer, and $\sigma$ is sigmoid function.

\subsubsection{LSTM neural network model for STLF}
%\FloatBarrier
The LSTM neural network includes an extra memory cell, which can overcome the gradient vanishing problem of recurrent neural network \citep{lstm_invent}. It has gain great popularity and is proven to be one of the most successful for load forecasting \citep{intro_16, lstm_apply2}.
%\begin{figure}[!ht]
%\hspace*{-0.3cm}
%\centering\includegraphics[scale=0.18]{LSTMcell2.jpg}\caption{The structure of LSTM cell}
%\label{fig:lstmcell}
%\end{figure}
%\FloatBarrier
\begin{figure}[!htb]
%\vspace*{-2.5cm}
\centering\includegraphics[scale=0.38]{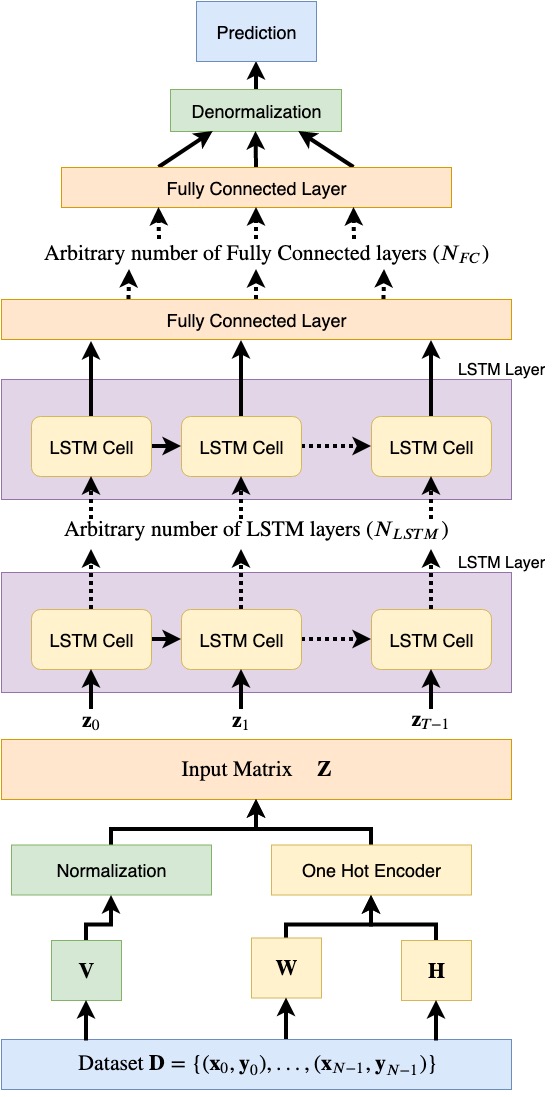}
\caption{LSTM forecasting model}
\label{fig:lstmgeneralmodel}
\end{figure}
%此图中的N改为N-1

The architecture of LSTM load forecasting model is shown in Fig. \ref{fig:lstmgeneralmodel}. The original dataset is denoted by ${D}=\{({ x}_0, {y}_0),...,({x}_{N-1}, { y}_{N-1})\}$, where $N$ is the number of samples in the dataset. Then input matrix ${Z}$ is concatenated by three vectors,
\iffalse
\begin{itemize}
    \item $N$ is the number of samples in the dataset.
    \item ${x}_i\in \mathcal{R}^{T}\ \forall i$ where $T$ is the number of look-back timestamps.
    \item ${y}_i\in \mathcal{R}^{T^{'}}\ \forall i$ where $T^{'}$ is the number of look-ahead timestamps.
    \item The model maximizes the probability of $P({y}_i|{x}_i)$ to give a prediction for ${y}_i$.
\end{itemize}
\fi

\begin{linenomath*}\begin{equation}
    {Z} = [{V}, { W}, { H}]\label{lstm_concat}
\end{equation}\end{linenomath*}
where ${V}$ is normalization vector of load dataset, ${W} $ is weekday indices vector, which is introduced to differentiate from Monday to Sunday, and ${ H} $ is holiday indices vector (i.e., holiday is 1, and non-holiday is 0). 

After that, input vector ${Z}$ is fed into arbitrary number of LSTM layers. And the output of the last LSTM layer is fed again into a fully connected neural network with $N_{FC}$ layers. The last layer of the fully connected neural network has dimension $\mathcal{R}^{T^{'}}$, in which each entry corresponds to forecast value of each timestamp. The final prediction is calculated by an additional de-normalization layer after the fully connected neural network.

\section{Experiments}\label{sec:results}
To verify the effectiveness of the proposed hybrid strategy for STLF, we tested the models on two datasets, one on the provincial level and one on the municipal level. In the experiments, each of the four inference models is fed with 7 days' hourly data and forecasts 1 day's hourly data ahead. Their outputs are then concatenated and fed to every ensemble model which gives a final prediction on the next 24 hour's hourly data. To be fair for comparison, the same ARIMA, NuSVR, ELM, and LSTM models are used throughout the experiment. Their parameters are tuned to be the best we can find. Detailed parameter selection process is described in Section \ref{ssec:submodel}. The outputs from the four selected models are then used for the training and testing of ensemble methods. The method we used for the training of WGTB is early stopping. Its effectiveness is verified in Section \ref{sssec:early_stopping}. And we ran a comprehensive grid search to determine the optimal hyperparameters. Finally, the results are compared and analyzed.

In the analysis process, we first compare the conventional hybrid models with single submodels to show the effectiveness of hybridizing both low-bias and low-variance submodels. We then compare the proposed hybrid model with conventional hybrid models to demonstrate that bias and variance are reduced concurrently under warm-start initialization and the hybrid boosting-bagging convergence. Additionally, we analyze the effectiveness of the integration of bagging in boosting in WGTB on the reduction of variance in Section \ref{sssec:early_stopping}. The section also proves empirically the feasibility of the early stopping method in the training of WGTB. The early stopping of WGTB effectively ensures that the practitioner can always get a definitive optimal error rate for each vector in hyperparameter space, which means that, unlike neural nets, the practitioners only need to run once for each hyperparameters combination. As a result, the early stopping greatly ease the hyperparameter tuning.

Three statistical indices (i.e., mean absolute percent error (MAPE), mean absolute error (MAE) and root mean square error (RMSE)) are utilized to analyze the forecasting effect.
\begin{linenomath*}\begin{equation}
\text{MAPE}=\frac{1}{N}\sum_{i=0}^{N-1}\frac{1}{T}\sum_{t=0}^{T-1}\left|\frac{y_t-y^{'}_t}{y_t}\right|\times 100\%
\label{eqn:metrics_mape}
\end{equation}\end{linenomath*}
\begin{linenomath*}\begin{equation}
    \text{MAE}=\frac{1}{N}\sum_{i=0}^{N-1}\frac{1}{T}\sum_{t=0}^{T-1}\left| y_t-y^{'}_t\right|\label{eqn:metrics_mae}\\
\end{equation}\end{linenomath*}
\begin{linenomath*}\begin{equation}
  \text{RMSE}=\frac{1}{N}\sum_{i=0}^{N-1}\sqrt{\frac{1}{T}\sum_{t=0}^{T-1}(y_t-y^{'}_t)^{2}}\label{eqn:metrics_rmse}
\end{equation}\end{linenomath*}
where $T$ is the number of load forecasting, $N$ is the number of test samples, $y_t$ is the observed value for the time period $t$, and $y^{'}_t$ is the forecasting value for the corresponding period.
\begin{table*}

\centering
\caption{Final selection of submodels\label{tab:submodel_selection}}
\begin{tabular}{ p{1.3cm} c c c c c c c c c } 
 \hline
 Submodel & One-hot Inputs & $p$ & $d$ & $q$ & Kernel & $\nu$ & $\#$ Neurons & Learning Rate [Epoch]\\
 \hline
 ARIMA & None & 1 & 1 & 1 & - & - & - & -\\
 NuSVR & None & - & - & - & Rbf & 0.1 & - & -\\
 ELM & Holiday & - & - & - & - &  & 1800 & -\\
 LSTM & Holiday, Weekday & - & - & - & - & - & 2$\times$128 & 0.001[100 epochs]+\\
 &&&&&&&&0.0001[130 epochs]\\
 \hline
\end{tabular}

\end{table*}

\begin{table*}

\centering
\caption{Comparison chart of one-hot inputs to LSTM in Experiment 1\label{tab:inputs}}
\begin{tabular}{ p{3.5cm} c c c c c c } 
 \hline
 One-hot Inputs & \multicolumn{3}{c}{Train} & \multicolumn{3}{c}{Test} \\ 
 \cline{2-7}
 & MAPE & MAE & RMSE & MAPE & MAE & RMSE\\
 \hline
 None & 2.163 & 1344.958 & 1607.524 & 1.606 & 1091.066 & 1307.625\\ 
 %\hline
 Holiday & 1.843 & 1156.246 & 1391.63623 & \textbf{1.217} & \textbf{824.463} & \textbf{1001.459}\\ 
 %\hline
 Holiday, Hour & 1.795 & 1121.056 & 1358.471 & 1.279 & 867.106 & 1059.715\\ 
 %\hline
 Weekday, Hour & 1.574 & 974.020 & 1182.820 & 1.300 & 887.606 & 1076.916\\ 
 %\hline
 Holiday, Weekday, Hour & \textbf{1.351} & \textbf{843.388} & \textbf{1037.643} & 1.284 & 874.699 & 1074.143\\
 %\hline
 \textbf{Holiday, Weekday} & \textbf{1.252} & \textbf{782.407} & \textbf{956.200} & \textbf{1.263} & \textbf{849.528} & \textbf{1028.526}\\
 \hline
\end{tabular}

\end{table*}
\subsection{Experiment on provincial data}
The dataset in Experiment 1 records the actual load of Jiangsu province from State Grid Corporation of China. The record spans from 1 o'clock on January 1, 2017 to 24 o'clock on December 31, 2017. It's based on one-hour basis (i.e., 24 data points each day). The first 300 days of the data set are used for training, and the rest are used for testing.
%\FloatBarrier
\subsubsection{Submodel Selection}
\label{ssec:submodel}
As illustrated is Section \ref{sec:overview}, four submodels are first used to forecast the load before ensemble. We present our selection mechanism for the individual submodels in the following subsections. And Table \ref{tab:submodel_selection} lists the final selection for submodels as well as their training parameters.

\begin{table*}
\centering
\caption{Evaluation indices of conventional hybrid models and individual submodels in Experiment 1}\label{tab:submodels}
\begin{tabular}{ p{2cm} c c c c c c }  \hline
 Algorithm & \multicolumn{3}{c}{Train} & \multicolumn{3}{c}{Test} \\ 
 \cline{2-7}
 & MAPE & MAE & RMSE & MAPE & MAE & RMSE\\
 \hline
 ARIMA & 100.033 & 63730.902 & 63881.906 & 8.256 & 5428.301 & 6578.890\\ 
 %\hline
 NuSVR & 2.628 & 1639.291 & 1943.286 & 1.951 & 1295.678 & 1537.575\\ 
 %\hline
 ELM & 1.456 & 920.775 & 1093.486 & 1.413 & 948.076 & 1128.245\\
 %\hline
 LSTM & 1.252 & 782.407 & 956.200 & 1.263 & 849.528 & 1028.526\\
 %\hline
 %{'alpha': 0.0001, 'l1_ratio': 1.0}
 \textbf{ElasticNet} & \textbf{1.231} & 769.381 & 940.674 & \textbf{1.229} & \textbf{826.002} & \textbf{999.726}\\
 %\hline
 %n_est:90 lr:0.050000 dep:3 found by grid search
 \textbf{SGTB} & \textbf{1.231} & \textbf{767.546} & \textbf{937.130} & 1.243 & 836.016 & 1011.522\\
 \hline
\end{tabular}
\end{table*}

\begin{figure*}
%\begin{figure}
\centering
\subfigure[Sample 1]{
\includegraphics[scale=0.29]{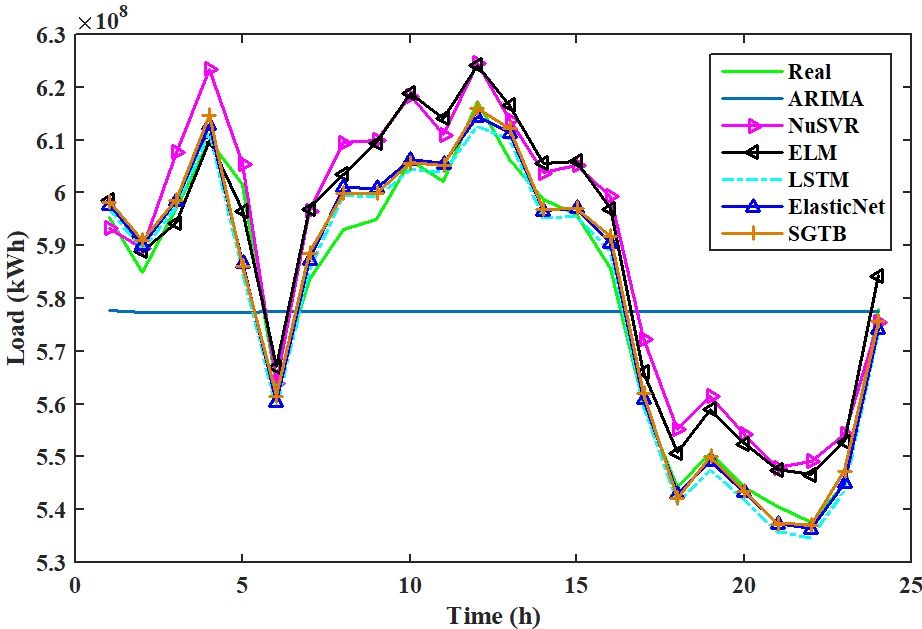}
}

\subfigure[Sample 2]{
\includegraphics[scale=0.29]{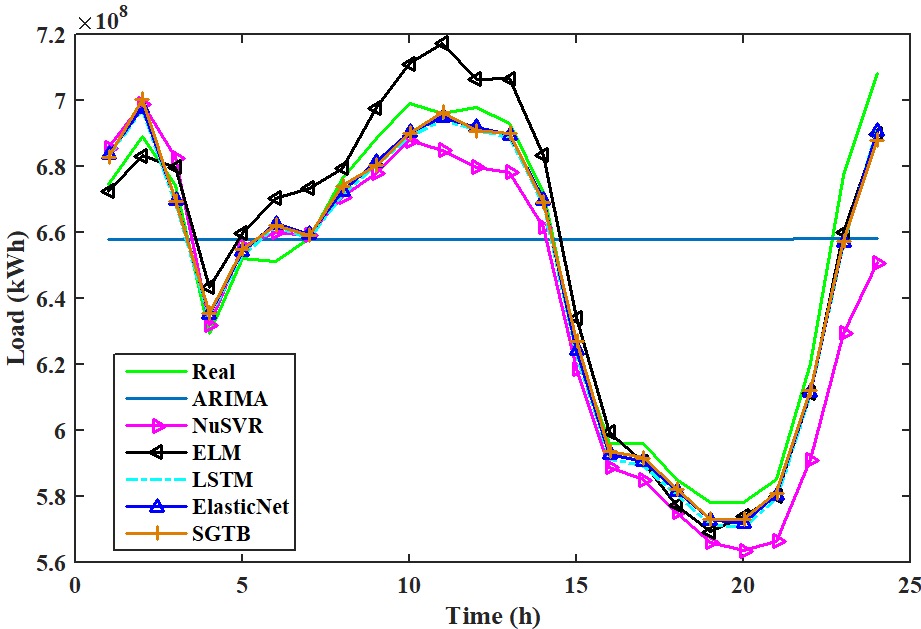}}
\caption{Forecasting results of conventional hybrid models and individual submodels in Experiment 1}
\label{fig:con ensem_single}
\end{figure*}

\begin{figure*}
%\begin{figure}[!htb]
\centering
\subfigure{
\includegraphics[scale=0.4]{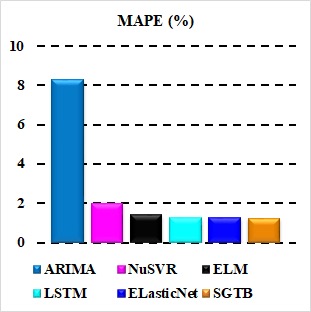}
}
\subfigure{
\includegraphics[scale=0.4]{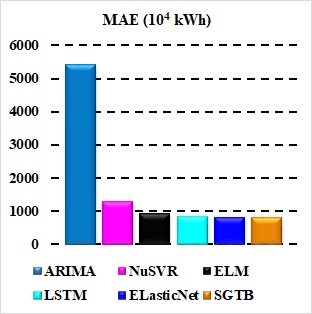}}
\subfigure{
\includegraphics[scale=0.4]{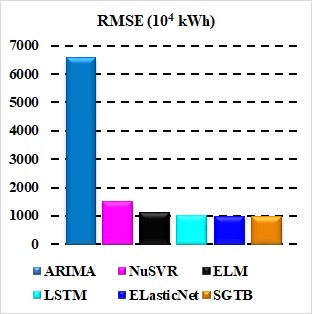}}
\caption{The comparison of conventional hybrid models and individual submodels in Experiment 1}
\label{fig:excel1}\end{figure*}
\begin{figure*}
\centering
\subfigure[Sample 1]{
\includegraphics[scale=0.29]{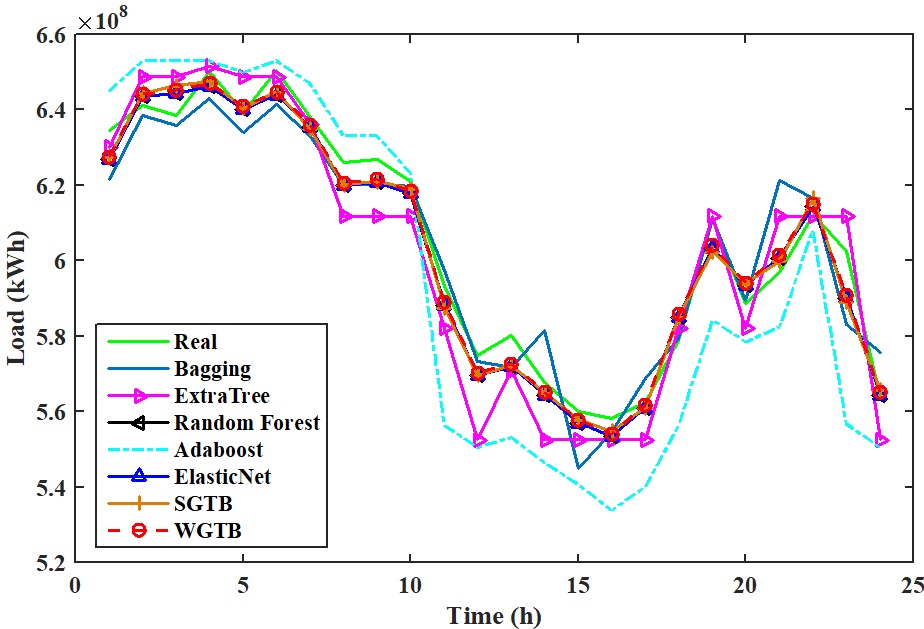}
}

\subfigure[Sample 2]{
\includegraphics[scale=0.29]{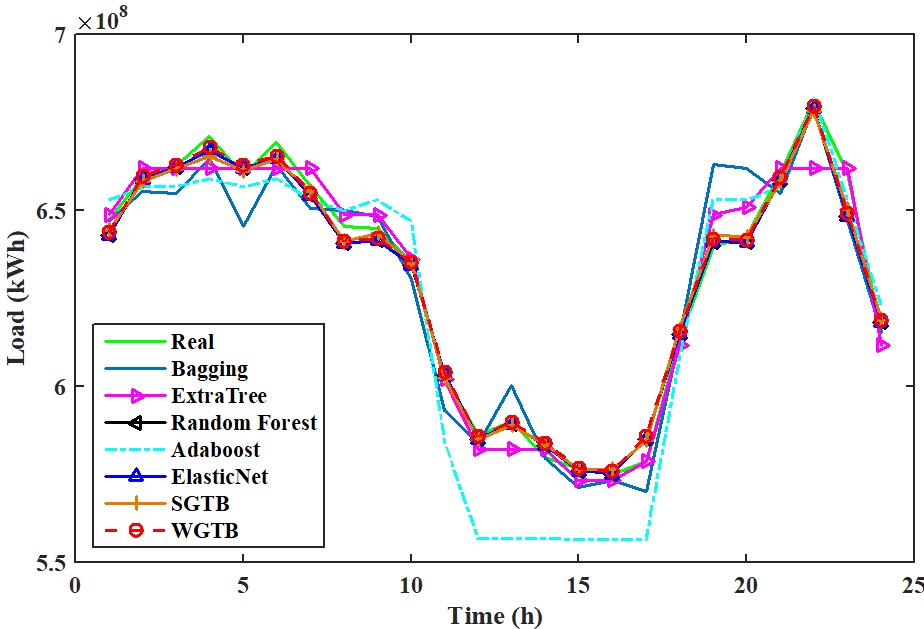}}
\caption{Forecasting results of proposed hybrid model and conventional hybrid models in Experiment 1}
\label{fig:proposed_con}
\end{figure*}

\begin{figure*}
\centering
\subfigure{
\includegraphics[scale=0.4]{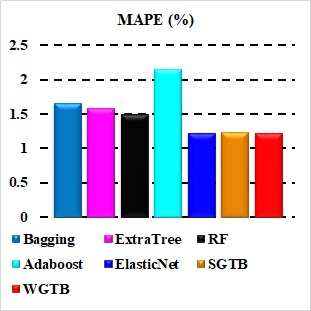}
}
\subfigure{
\includegraphics[scale=0.4]{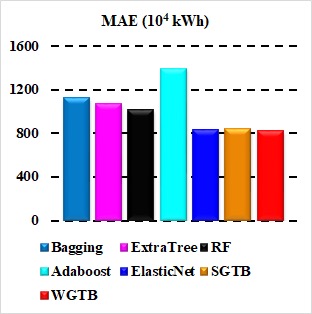}}
\subfigure{
\includegraphics[scale=0.4]{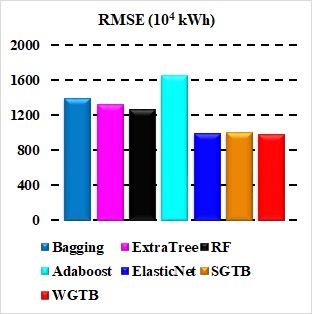}}
\caption{The comparison of  proposed hybrid model and conventional hybrid models in Experiment 1}
\label{fig:excel2}\end{figure*}
\begin{table*}
\centering
\caption{Evaluation indices of proposed hybrid model and conventional hybrid models in Experiment 1\label{tab:proposed_con}}
\begin{tabular}{ p{3cm} c c c c c c } 
 \hline
 Algorithm & \multicolumn{3}{c}{Train} & \multicolumn{3}{c}{Test} \\ 
 \cline{2-7}
 & MAPE & MAE & RMSE & MAPE & MAE & RMSE\\
 \hline
 %n_est:5
 Bagging & \textbf{0.313} & \textbf{195.067} & \textbf{319.824} & 1.665 & 1125.390 & 1400.803\\
 %\hline
 %max_depth=10
 ExtraTree & 1.591 & 984.398 & 1225.495 & 1.594 & 1068.562 & 1329.800\\
 %\hline
 %depth:10 n:100
 Random Forest & \textbf{1.057} & \textbf{661.133} & \textbf{813.568} & 1.496 & 1016.035 & 1268.435\\
 %\hline
 %n_estimators = 200, loss='linear'
 Adaboost & 2.450 & 1469.618 & 1770.931 & 2.160 & 1396.044 & 1665.956\\
 %\hline
 %{'alpha': 0.0001, 'l1_ratio': 1.0}
 ElasticNet & 1.231 & 769.381 & 940.674 & 1.229 & 826.002 & 999.726\\
 %\hline
 %n_est:90 lr:0.050000 dep:3 found by grid search
 SGTB & 1.231 & 767.546 & 937.130 & 1.243 & 836.016 & 1011.522\\
 %\hline
 %ITER 460 lr:0.01 dep:3 random_split:0.5
 \textbf{WGTB} & \textbf{1.219} & \textbf{763.543} & \textbf{932.645} & \textbf{1.218} & \textbf{817.112} & \textbf{988.924} \\
 \hline
\end{tabular}

\end{table*}

%\afterpage{\newpage}
(1) ARIMA

The key of ARIMA model is to determine the three parameters $p$, $d$, $q$. The autocorrelation function  and partial autocorrelation function are utilized to choose proper values. Through the corresponding analysis, the model is selected as ARIMA(1,1,1).  
%Three orders (i.e., 0, 1 and 2) of $d$ of training set are drawn respectively in Figs. \ref{fig:Autocor0}-\ref{fig:Autocor2}. It can be seen from the figures that first order and second order both share low variance. But $d$ is chosen to be 1 rather than 2 because of standard deviation. Standard deviation of first order is 2150.006, which is lower than 2334.848 of second order. And AR term and MA term should be cutoff value in PACF and ACF respectively. So $p$ and $q$ are chosen to be 1. Finally ARIMA(1,1,1) is selected as the submodel.
%\begin{figure}[H]
%\vspace{-0.2cm}
%\centering\includegraphics[scale=0.27]{Autocorrelation0th@600.jpg}\caption{ACF and PACF of order 0}\label{fig:Autocor0}

%\includegraphics[scale=0.27]{Autocorrelation1st@600.jpg}\caption{ACF and PACF of order 1}\label{fig:Autocor1}

%\includegraphics[scale=0.27]{Autocorrelation2nd@600.jpg}\caption{ACF and PACF of order 2}\label{fig:Autocor2}
%\end{figure}

%\FloatBarrier
(2) NuSVR

In the experiment on our dataset, the RMSE, MAE and MAPE of SVR are twice to three times as that of NuSVR. Thus, NuSVR is selected to represent SVR family in the proposed model. The Rbf kernal is used, because it outperforms other kernels tested, including polynomial, exponential and linear. And the parameter $\nu$ is tested to be 0.1.

(3) ELM

 The one-hot inputs of ELM are determined in a similar method as LSTM fully described in the next paragraph. And for each set of inputs, the number of neurons is determined by the offset where training set error is similar to test set error. This selection strategy comes from the fact that more neurons lead to overfitting while less neurons lead to underfitting. In our final selection, the number of neurons in the hidden layer is 1800, and its inputs are holiday and preceding 168 hourly load data.
 
(4) LSTM

The LSTM model consisting of two-layer LSTM with 128 hidden neurons and one-hidden-layer fully connected is the most suitable model in testing. The experiment data for one-hot input selection is shown in Table \ref{tab:inputs}. The units of MAE and RMSE in Tables \ref{tab:inputs}-\ref{tab:proposed_con2} are $10^4$ kWh. The best two numeric values are highlighted in bold. It is easy to find that holiday indices improve the accuracy, weekday indices alleviate overfitting and underfitting issues, and hour indices generally reduce the performance of prediction. So our choice of the one-hot inputs of LSTM are holiday and weekday.

\subsubsection{Conventional hybrid models versus single submodels}
The individual regression models (ARIMA, NuSVR, ELM, and LSTM) with the hyperparameters specified in Section \ref{ssec:submodel} are compared with the hybrid strategies based on conventional ensemble models (ElasticNet and SGTB). Fig. \ref{fig:con ensem_single} visualizes the predictions from conventional hybrid models versus single submodels in two arbitrary samples. The evaluation indices calculated by our dataset are listed in Table \ref{tab:submodels}, out of which sixty-five samples compose the test set. Fig. \ref{fig:excel1} provides the corresponding histograms. As shown in Table \ref{tab:submodels} and Fig. \ref{fig:excel1}, LSTM is manifestly the best single estimator for STLF. Still, both ensemble models, ElasticNet and SGTB, outstrip it for all indices tested. MAPE, MAE, RMSE of ElasticNet are 2.692\%, 2.769\% and 2.800\% lower than those of LSTM in the test set. And MAPE, MAE, RMSE of SGTB are 1.584\%, 1.591\% and 1.653\% lower than those of LSTM in the test set. Accordingly the results show that hybrid method is effective in improving the accuracy by taking both low-bias and low-variance models into account in the selection of inference submodels. This part corresponds to the first novelty point of the hybrid strategy.
%\FloatBarrier

\subsubsection{The proposed hybrid model versus conventional hybrid models}
The proposed hybrid model based on WGTB is compared with conventional hybrid models based on ensemble models (Bagging, ExtratTree, Random forest, Adaboost, ElasticNet, and SGTB). Fig. \ref{fig:proposed_con} shows the prediction result, and evaluation indices are listed in Table \ref{tab:proposed_con}. Fig. \ref{fig:excel2} provides the corresponding histograms. It is easy to find that the best two traditional ensemble methods are ElasticNet and SGTB, whereas other traditional ensemble methods have higher error rates. But the proposed WGTB further beats both ElasticNet and SGTB in all aspects. It does not merely have lower error rates in three evaluation indices and for the training set and test set, but it also has the smallest generalization errors between the training set and test set, especially in MAPE. More specifically in terms of the first aspect on error rates, MAPE, MAE, RMSE of WGTB are 0.895\%, 1.076\%, and 1.080\% lower than those of ElasticNet, and 2.011\%, 2.261\% and 2.234\% lower than those of SGTB in the test set. And as for the second aspect on the generalization error, generalization error of WGTB are 0.001\%, 535690 kWh and 562790 kWh in the order of MAPE, MAE and RMSE. By contrast, generalization errors of ElasticNet are 0.002\%, 566210 kWh and 590520 kWh and generalization errors of SGTB are 0.012\%, 684700 kWh and 743920 kWh, both in the same order as the preceding sentence. Since both error rate and generalization errors are reduced, according to the learning theory, bias and variance must have been reduced concurrently. This part corresponds to the second and the third novelty points of the hybrid strategy.

\subsubsection{Variance analysis of the integration of bagging in boosting \& Early stopping}\label{sssec:early_stopping}
\begin{figure*}
\includegraphics[scale=0.04]{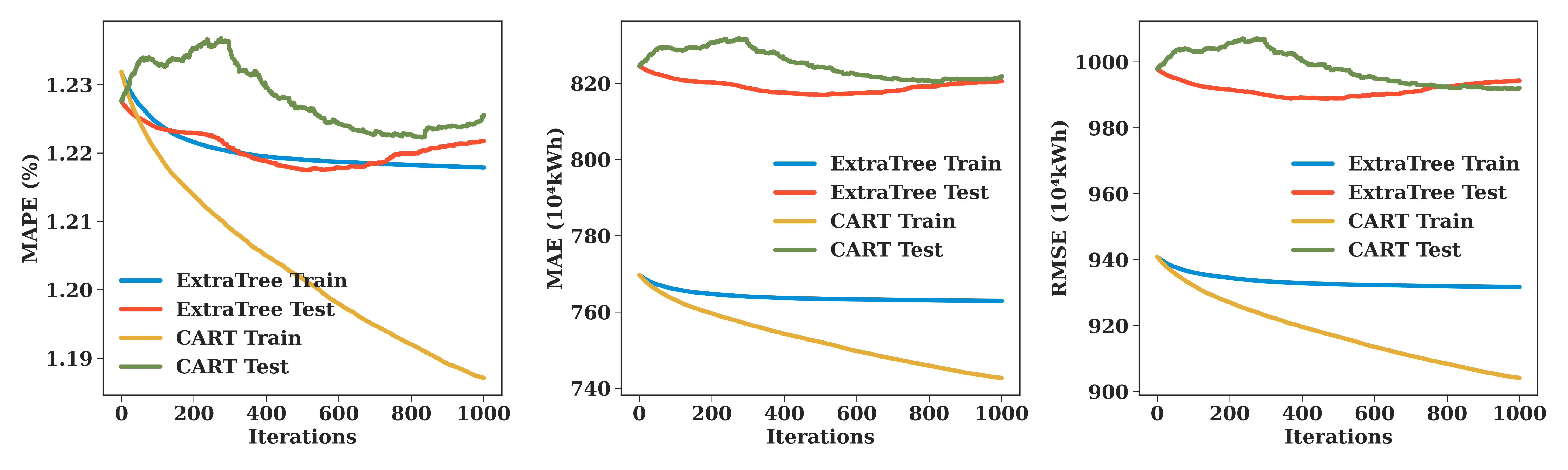}
\caption{ExtraTree versus decision tree in Experiment 1}\label{fig:CARTvsExtratree}\end{figure*}

This subsection verifies the variance analysis of the use of bagging in the boosting model. We cannot directly compute the variance since base tree estimators are constructed online via an iterative algorithm. The underlying ground-truth distribution changes with the number of iterations and depends on previous iterations. Therefore, we take another route and test the overfitting of boosting. Boosting is known to lower the bias but can lead to overfitting when the total variance is too large. To see if bagging reduces the variance of boosting, we compare WGTB with a newly constructed ensemble model. The new model is the same as WGTB except for its base tree estimator is modeled by decision tree (i.e., CART) instead of ExtraTree. Fig. \ref{fig:CARTvsExtratree} plots the training errors in the first 1000 iterations when either CART or ExtraTree is used in boosting. For the CART-based model, whereas its training error quickly converges, its testing error diverges in the first 300 iterations before converging and reaching its minimum at around 900 iterations. This phenomenon reflects the high variance brought by decision trees. And the high variance harms the model in three ways. Firstly, it causes a severe overfitting issue. Secondly, even with overfitting, its testing error rate is still higher than that of the proposed ExtraTree-based model. Thirdly, it takes more iterations to converge.

On the contrary, the low variance of ExtraTree leads to a stable convergence. The randomization and bagging allow the boosting to be more stable and suffer less from variance and overfitting. Also, note the trend that the testing errors drop together with training errors before it reaches its optimum and bounces back in testing errors. The trend suggests the early stopping method can be used to determine the optimal generalization performance.%Thus the early stopping method can be used to determine the optimal generalization performance.%The trend suggests an early stopping method to be used in the training of WGTB.
\begin{table*}
\centering
\caption{Evaluation indices of conventional hybrid models and individual submodels  in Experiment 2}\label{tab:ensem_single2}
\begin{tabular}{ p{2cm} c c c c c c }  \hline
 Algorithm & \multicolumn{3}{c}{Train} & \multicolumn{3}{c}{Test} \\ 
 \cline{2-7}
 & MAPE & MAE & RMSE & MAPE & MAE & RMSE\\
 \hline
 ARIMA & 100.026 & 2040.518 & 2053.471 & 9.525 & 198.958 & 233.521\\ 
 %\hline
 NuSVR & 3.237 & 65.557 & 76.511 & 2.987 & 65.015 & 79.049\\ 
 %\hline
 ELM & 2.101 & 43.153 & 50.844 & 1.928 & 42.276 & 50.484\\
 %\hline
 LSTM & 1.789 & 36.291 & 43.052 & 1.784 & 38.829 & 47.204\\
 %\hline
 %{'alpha': 0.0001, 'l1_ratio': 1.0}
 \textbf{ElasticNet} & 1.628 & 33.352 & 39.853 & 1.686 & 37.084 & \textbf{45.361}\\
 %\hline
 %n_est:90 lr:0.050000 dep:3 found by grid search
 \textbf{SGTB} & \textbf{1.614} & \textbf{32.962} & \textbf{39.484} & \textbf{1.685} & \textbf{37.073} & 45.611\\
 \hline
\end{tabular}
\end{table*}
\begin{table*}
\centering
\caption{Evaluation indices of proposed hybrid model and conventional hybrid models in Experiment 2}\label{tab:proposed_con2}
\begin{tabular}{ p{3cm} c c c c c c } 
 \hline
 Algorithm & \multicolumn{3}{c}{Train} & \multicolumn{3}{c}{Test} \\ 
 \cline{2-7}
 & MAPE & MAE & RMSE & MAPE & MAE & RMSE\\
 \hline
 %n_est:5
 Bagging & \textbf{0.449} & \textbf{9.107} & \textbf{11.993} & 1.991 & 43.191 & 53.019\\
 %\hline
 %max_depth=10
 ExtraTree & 1.637 & 33.570 & 41.018 & 1.775 & 38.951 & 48.060\\
 %\hline
 %depth:10 n:100
 Random Forest & 2.035 & 41.370 & 49.879 & 2.044 & 44.773 & 54.425\\
 %\hline
 %n_estimators = 200, loss='linear'
 Adaboost & 2.184 & 44.786 & 54.100 & 2.297 & 50.479 & 61.522\\
 %\hline
 %{'alpha': 0.0001, 'l1_ratio': 1.0}
 ElasticNet & 1.628 & 33.352 & 39.853 & 1.686 & 37.084 & 45.361\\
 %\hline
 %n_est:90 lr:0.050000 dep:3 found by grid search
 SGTB & 1.614 & \textbf{32.962} & 39.484 & 1.685 & 37.073 & 45.611\\
 %\hline
 %ITER 460 lr:0.01 dep:3 random_split:0.5
 \textbf{WGTB} & \textbf{1.612} & \textbf{32.998} & \textbf{39.480} & \textbf{1.665} & \textbf{36.540} & \textbf{44.650} \\
 \hline
\end{tabular}
\end{table*}

\subsection{Experiment on municipal data}
We have conducted an additional experiment on the municipal electrical load data to verify the proposed model's generalization performance. The dataset in Experiment 2 records the actual load of Nanjing city from State Grid Corporation of China. The record spans from 1 o'clock on January 1, 2014 to 24 o'clock on December 31, 2014 with hourly resolution. The first 300 days of the data set are split for training, and the rest are reserved for testing. The experiment process is kept the same as Experiment 1.
\subsubsection{Conventional hybrid models versus single submodels}
\begin{figure*}[!htb]
%\begin{figure}
\centering
\includegraphics[scale=0.29]{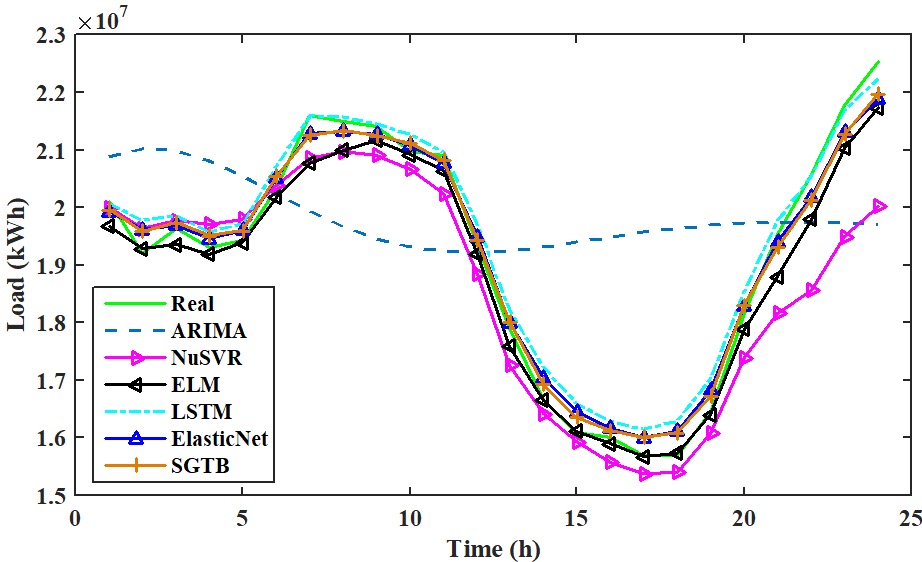}
\caption{Forecasting results of conventional hybrid models and individual submodels in Experiment 2}
\label{fig:ensem_single2}
\end{figure*}

 The conventional hybrid models and individual submodels are compared in Experiment 2. Fig. \ref{fig:ensem_single2} shows an arbitrary sample of their forecasting results, and Table \ref{tab:ensem_single2} further lists their evaluation indices. The table shows that the ensemble models ElasticNet and SGTB have better performance than the best submodel LSTM. The MAPE, MAE, RMSE of ElasticNet are 5.493\%, 4.494\%, and 3.904\% lower than those of LSTM in the test set. Moreover, MAPE, MAE, RMSE of SGTB are 5.549\%, 4.452\%, and 3.375\% lower than those of LSTM in the test set. This verifies the first novelty point and the generalization performance of the hybrid strategy.
\subsubsection{The proposed hybrid model versus conventional hybrid models}
\begin{figure*}
%\begin{figure}
\centering
\includegraphics[scale=0.29]{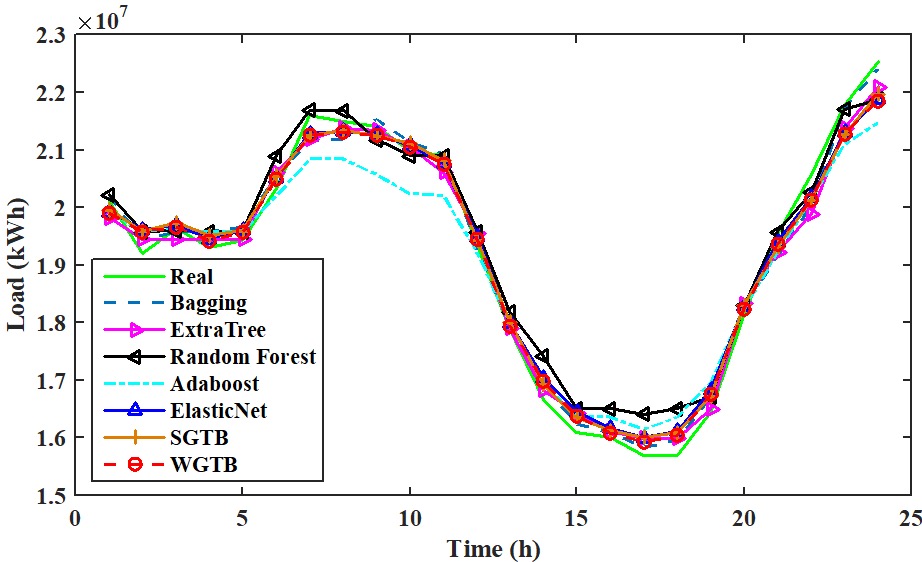}
\caption{Forecasting results of proposed hybrid model and conventional hybrid models in Experiment 2}
\label{fig:proposed_con2}
\end{figure*}

 The forecasting results of the proposed hybrid model and conventional hybrid models in Experiment 2 are shown in Fig. \ref{fig:proposed_con2} and Table \ref{tab:proposed_con2}. MAPE, MAE, RMSE of WGTB are 1.246\%, 1.467\%, and 1.567\% lower than those of ElasticNet in the test set, and they are 1.187\%, 1.438\%, and 2.107\% lower than those of SGTB in the test set. The generalization errors of WGTB are 0.052\%, 35420 kWh, and 51700 kWh in the order of MAPE, MAE and RMSE. By contrast, the generalization errors of ElasticNet are 0.058\%, 37320 kWh, and 55080 kWh and the generalization errors of SGTB are 0.071\%, 41110 kWh, and 61270 kWh, both in the same order as the preceding sentence. The WGTB achieves the lowest error rates and has better generalization errors. So bias and variance must have been reduced concurrently. This result verifies the second and the third novelty points and the generalization performance of WGTB.


\section{Conclusion}\label{sec:conclusion}
 A novel hybrid strategy based on WGTB is proposed to forecast the short-term load. The proposed strategy outperforms existing strategies by taking bias-variance trade-off into account in two ways. First, its inference submodels range from low-bias LSTM model to low-variance ARIMA model. Second, a novel ensemble model WGTB is proposed to lower bias and variance concurrently. The two objectives are achieved separately by a warm-start from ElasticNet and a hybrid of Boosting and Bagging during convergence. Experiments have been conducted to prove the effectiveness of our proposed strategy in the improvement of accuracy. The three statistical indices MAPE, MAE and RMSE of the proposed WGTB model are 3.563\%, 3.816\%, and 3.85\% lower than those of the best individual submodel LSTM, 2.011\%, 2.261\%, and 2.23\% lower than the best conventional hybrid model based on SGTB in the provincial electrical load dataset. An additional experiment has been conducted on the municipal load dataset to further verify the generalization performance. Nonetheless, there is still room for improvement in terms of speed. In future work, we will consider speeding up the algorithm by making a histogram-based warm-start gradient tree boosting model.

\section*{Declaration of Competing Interest}
The authors declare that they have no known competing financial interests.

\section*{Data Availability Statement}
The data that support the findings of this study are available from the corresponding author
upon reasonable request.

\section*{Author's Contributions}
Yuexin and Jiahong contributed equally to this work.

\begin{acknowledgments}
This work was supported by China Scholarship Council.
\end{acknowledgments}

\phantomsection

\end{document}